\documentclass[conference]{IEEEtran}
\IEEEoverridecommandlockouts

\usepackage{cite}
\usepackage{amsmath,amssymb,amsfonts}
\usepackage{graphicx}
\usepackage{textcomp}
\usepackage{xcolor}
\def\BibTeX{{\rm B\kern-.05em{\sc i\kern-.025em b}\kern-.08em
    T\kern-.1667em\lower.7ex\hbox{E}\kern-.125emX}}

\usepackage[utf8]{inputenc}
\usepackage[T1]{fontenc}
\usepackage{booktabs}
\usepackage{multirow}
\usepackage{algorithm}
\usepackage{algpseudocode}
\usepackage{microtype}
\usepackage[draft,bookmarks=false]{hyperref}
\hypersetup{draft=true, bookmarks=false}
\usepackage{enumitem}
\usepackage{xcolor}

\begin{document}

\title{\textsc{QuantFL}: Sustainable Federated Learning for Edge IoT via Pre-Trained Model Quantisation
}

\author{
\IEEEauthorblockN{Charuka Herath\IEEEauthorrefmark{1}, Yogachandran Rahulamathavan\IEEEauthorrefmark{2}, Varuna De Silva\IEEEauthorrefmark{3}, Sangarapillai Lambotharan\IEEEauthorrefmark{3}}
\IEEEauthorblockA{\IEEEauthorrefmark{1}Institute of Digital Technologies, Loughborough University, UK\\
Email: \{c.herath, y.rahulamathavan, v.d.de-silva, s.lambotharan\}@lboro.ac.uk}
}

\maketitle

\begin{abstract}
Federated Learning (FL) enables privacy-preserving intelligence on Internet of Things (IoT) devices but incurs a significant carbon footprint due to the high energy cost of frequent uplink transmission. While pre-trained models are increasingly available on edge devices, their potential to reduce the energy overhead of fine-tuning remains underexplored. In this work, we propose \textsc{QuantFL}, a sustainable FL framework that leverages pre-trained initialisation to enable aggressive, computationally lightweight quantisation. We demonstrate that pre-training naturally concentrates update statistics, allowing us to use memory-efficient bucket quantisation without the energy-intensive overhead of complex error-feedback mechanisms. On MNIST and CIFAR-100, \textsc{QuantFL} reduces total communication by 40\% ($\simeq40\%$ total-bit reduction with full-precision downlink; $\geq80\%$ on uplink or when downlink is quantised) while matching or exceeding uncompressed baselines under strict bandwidth budgets; BU attains 89.00\% (MNIST) and 66.89\% (CIFAR-100) test accuracy with orders of magnitude fewer bits.  We also account for uplink and downlink costs and provide ablations on quantisation levels and initialisation. \textsc{QuantFL} delivers a practical, "green" recipe for scalable training on battery-constrained IoT networks.
\end{abstract}

\begin{IEEEkeywords}
Sustainable Federated Learning, Model Quantisation, Signal Processing, Communication-Efficiency, Internet of Things
\end{IEEEkeywords}

\section{Introduction}
\label{sec:intro}

Federated Learning (FL) has emerged as a powerful paradigm for decentralised machine learning, enabling multiple devices or organisations to collaboratively train models without sharing raw data~\cite{FedAvg2017}. This preserves data privacy and security while enabling knowledge sharing across clients \cite{DSFL}. However, despite its advantages, FL faces critical challenges in real-world deployments, such as the high communication overhead caused by transmitting large model updates or gradients to a central server. This issue becomes particularly acute in bandwidth-limited environments, such as edge devices or mobile networks, where communication cost dominates overall training efficiency.

One promising strategy to reduce communication rounds is to initialise training with pre-trained models~\cite{pre-trained, PPcontAuthFL2022}. Such models provide rich feature representations and often require fewer local updates to converge, making them attractive for FL settings. However, pre-trained models are typically large and overparameterised, and transmitting them (or their updates) still incurs high communication costs unless compression is applied.

Motivated by classical signal processing principles, we explore quantisation as a solution to this bottleneck. In this context, model updates in FL can be viewed as high-dimensional signals, and structured quantisation can reduce their bit-length without significantly affecting learning performance. By compressing these updates using signal-inspired quantisation techniques, we aim to enable scalable and efficient FL systems.

In this work, we propose \textsc{QuantFL}, a communication-efficient FL framework that combines partial model pre-training with structured quantisation. Specifically, \textsc{QuantFL} applies two bucket-based quantisation schemes: bucket-uniform (BU), which partitions update values into equal-width bins~\cite{GershoGray1992}, and bucket-quantile (BQ), which adapts bins based on empirical data distribution~\cite{AdpatDeep2017}. Then we benchmark them against a stochastic baseline, QSGD~\cite{QSGD} and Baseline FedAvg \cite{FedAvg2017}. Unlike previous works that focus primarily on quantising gradients during training, \textsc{QuantFL} compresses full pre-trained model updates, bridging a gap between model reuse and communication-efficient FL.

\subsection{Motivation}

In cross-device FL, uplink bandwidth is scarce: every round, each selected client must transmit a large model update. Starting from a pre-trained model changes the statistics of these updates: they become smaller and less dispersed across parameters. Intuitively, pre-training puts the model in a good region, so local steps are modest. This makes simple scalar quantisation far more effective: with fewer bits per parameter, we reach the same accuracy, so total communication drops sharply.
Our approach: \textsc{QuantFL} takes a pre-trained initialisation and quantises each client’s update every round, sending compact indices plus occasionally refreshed side information. In practice, this yields order-of-magnitude uplink savings while matching uncompressed accuracy, even under non-independent and identically distributed (non-IID) data. In practical application scenarios in distributed settings, devices increasingly have access to pre-trained backbones; \textsc{QuantFL} shows how to turn that into tangible communication savings during training.

\subsection{Contributions}
This paper introduces \textsc{QuantFL}, a communication-efficient FL framework that couples pre-trained model initialisation with range-aware bucketed scalar quantisation of client updates. Our key contributions are:
\begin{enumerate}
    \item Pre-training-aware compression - We empirically show pre-training concentrates $\Delta w$ (smaller range/variance), enabling fewer bits for the same distortion.
    \item Simple, deployable quantizers. Range-adaptive BU/BQ with mid-point decoding, per-layer boundaries, and explicit bit accounting for uplink/downlink.
    \item Robustness under heterogeneity. BU remains stable under non-IID; we include and Dirichlet-$\alpha$ sweeps.
    \item Communication–accuracy Pareto gains. $\geq98\%$ bit reduction at parity/better accuracy across datasets and lower loss and communication cost reduction under tight bandwidth budgets.
\end{enumerate}


\begin{table}[h]
\centering
\caption{List of Notations}
\vspace{-0.2cm}
\begin{tabular}{l l}
\hline
\textbf{Symbol} & \textbf{Description} \\
\hline
$N, \mathcal{S}_k$ & Total clients, Selected subset at round $k$ \\
$w^k, w_i^k$ & Global model, Local model of client $i$ \\
$\Delta w$ & Model update ($w_i^k - w^k$) \\
$L_l, \mathcal{B}_l$ & Quantization levels, Codebook boundaries for layer $l$ \\
$T_l$,$\alpha$ & Codebook refresh period, Dirichlet-sweeps\\
\hline
\vspace{-0.5cm}
\end{tabular}
\end{table}

\section{Related Work}
\label{sec: relate work}
Communication has long been the principal bottleneck in cross-device FL due to uplink constraints and frequent synchronisation. FedAvg~\cite{FedAvg2017} reduces the \emph{number} of rounds via local computation, but still transmits full-precision parameters each round.

Stochastic/bias-controlled quantisers such as QSGD~\cite{QSGD} and TernGrad~\cite{TernGrad} lower bit-widths while preserving convergence guarantees. Unified analyses of compressed optimisation in FL characterise the role of unbiasedness, variance, and error-feedback~\cite{pmlrcomp}, and adaptive compression schemes further tune the communication–accuracy trade-off during training~\cite{reddi2021adaptive}. Most of these works target online gradients (per mini-batch) rather than full per-round model updates.

Update sparsification (e.g., top-$k$) and sign-based compressors are widely used to cut uplink bits; periodic aggregation and momentum correction improve their stability in heterogeneous settings (e.g., FedPAQ~\cite{fedPaq}; see also SparseFL~\cite{SparseFL} and OCTAV~\cite{OCTAV}). These methods typically produce biased updates and often require explicit error-feedback to avoid accuracy degradation under non-IID data.

\begin{figure}
    \centering
    \includegraphics[width=0.8\linewidth]{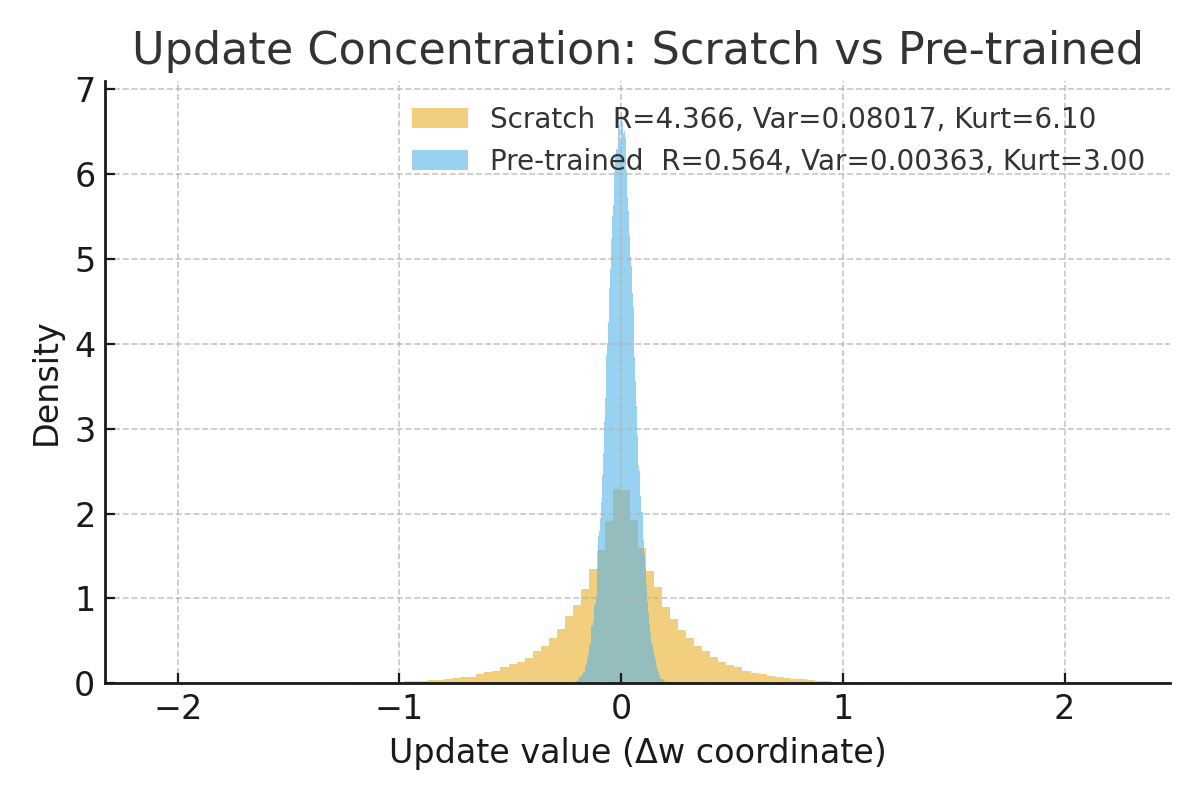}
    \vspace{-0.5cm}
    \caption{Update concentration with pre-training vs. training from scratch (simulated illustration). Pre-trained updates exhibit substantially smaller dynamic range R and variance, and reduced kurtosis (shorter tails), enabling lower quantisation error at fixed L.}
    \vspace{-0.5cm}
    \label{fig:update_concentration}
\end{figure}

FL optimisation often relies on complex compressors with Error Feedback (EF), where clients store and accumulate quantisation errors to correct future updates. While theoretically powerful, EF increases the local memory footprint and computational logic required on resource-constrained Internet of Things (IoT) devices, contributing to higher battery drain. We propose that pre-training serves as a cleaner, more energy-efficient alternative to EF. \cite{challanges_FL} By initialising from a pre-trained state, update variance is naturally minimised (as shown in Fig. \ref{fig:update_concentration}), enabling \textsc{QuantFL} to achieve high-fidelity compression using simple, memory-less scalar quantisation (BU/BQ). This eliminates the need for residual error storage, offering a "greener" trade-off that maintains accuracy while minimising the computational and energy burden on the edge device.

Matrix-structured compressors approximate updates with low-rank or sketch-based representations \cite{PowerSGD}. They are effective on large dense layers but introduce additional decomposition cost and hyperparameters (rank, sketch size), and are orthogonal to the scalar bucket approach we pursue. Beyond per-round compression, systems reduce the \emph{frequency} of communication (local steps, partial participation)~\cite{fedPaq}, and adapt compressor strength over time~\cite{reddi2021adaptive}. \textsc{QuantFL} is complementary: we keep standard round scheduling and reduce bits per round.

Post-training quantisation (PTQ) and quantisation-aware training (QAT) reduce model size and compute for deployment~\cite{surveyquant2021, AdpatDeep2017}, including on large backbones and ViTs~\cite{li2022ptq4vit}. These methods primarily target \emph{inference} efficiency; they neither account for per-round communication nor exploit pre-training to compress \emph{training-time} updates.

Warm starts and representation learning from pre-trained models accelerate convergence and improve personalisation in FL~\cite{augFL2025, PPcontAuthFL2022, pre-trained}. However, these lines of work do not specifically address how pre-training changes update statistics and can be leveraged to increase compression without harming accuracy.

Despite extensive research on gradient compression, sparsification, low-rank/sketching, and inference quantisation, there is limited exploration of pre-training–aware update quantisation in FL. \textsc{QuantFL} focuses on a simple, deployable scalar bucket quantiser applied to per-trained model updates every round, with explicit accounting of index bits and boundary refresh. Our key observation, that pre-training concentrates updates (smaller range/variance), connects classical scalar quantisation principles to modern FL: a reduced dynamic range yields lower distortion for a fixed number of levels, or equivalently, the same distortion at fewer bits. This pre-training aware perspective complements prior compressors and explains the robust communication savings we observe under non-IID data.


\begin{figure}[t]
    \centering
    \includegraphics[width=1\linewidth]{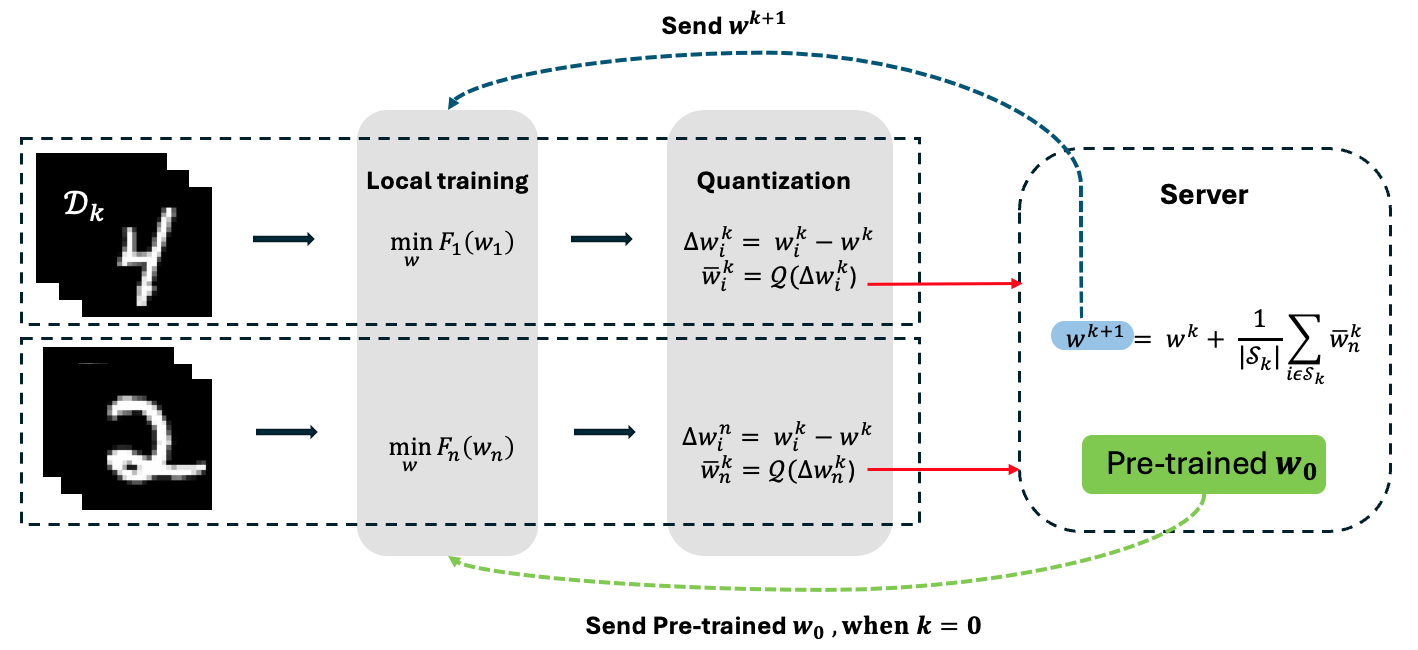}
    \caption{\textbf{\textsc{QuantFL} pipeline.} Each client quantises its per-layer update using \emph{BU} (equal-width) or \emph{BQ} (equal-mass) buckets and sends only \emph{indices}; \emph{boundaries} are refreshed infrequently. The server applies \emph{mid-point decoding} and aggregates. Pre-training makes updates narrowly distributed, enabling aggressive compression with little error.}
    \vspace{-0.5cm}
    \label{fig:system_model}
\end{figure}

\section{System Model and Problem Formulation}
\label{sec: system model}

We consider a classical FL setup consisting of a central server and $N$ distributed clients, indexed by $i \in \{1, \dots, N\}$. The server coordinates $K$ communication rounds to collaboratively train a global model $w \in \mathbb{R}^d$ without directly accessing clients' local data.

At the beginning of training, the server initialises the global model $w^0$ with a partially pre-trained model.  This initialisation captures general feature representations, enabling faster convergence. However, even partially pre-trained models are often large, motivating the need for communication-efficient strategies.

In each communication round $k \in \{0, \dots, K-1\}$: The server distributes the current global model $w^k$ to a subset of selected clients $\mathcal{S}_k$. Each client $i \in \mathcal{S}_k$ locally updates the model using its private dataset $\mathcal{D}_i$, by optimizing a local objective function:
    \begin{equation}
        \min_{w} F_i(w) = \frac{1}{|\mathcal{D}_i|} \sum_{(x_j, y_j) \in \mathcal{D}_i} \ell(w; x_j, y_j)
    \end{equation}
    where $\ell(w; x_j, y_j)$ is the loss function evaluated on data sample $(x_j, y_j)$.
After local training, each client obtains a local model $w_i^k$ and computes the model update:
    \begin{equation}
        \Delta w_i^k = w_i^k - w^k
    \end{equation}

\subsection{Communication Cost Model}

In standard FL without compression, the communication cost per client per round is:
\begin{equation}
    C_{\text{baseline}} = 32 \times d \quad \text{(bits)}
\label{eq:baseline}
\end{equation}
assuming 32-bit floating point representation per parameter.

Equation. \ref{eq:baseline} gives the uplink cost. Our tables report \emph{uplink+downlink} totals, hence $C_{\text{baseline,total}}=64\,d$ bits per client per round (32$d$ uplink + 32$d$ downlink).

With quantisation, each client compresses its local update before transmission. The communication cost per client is modelled as:

\begin{equation}
C_{\text{uplink}} \;=\; \sum_{\ell} \Big( d_\ell \cdot \lceil \log_2 L_\ell \rceil \;+\; \frac{b \cdot L_\ell}{T_\ell} \Big).
\label{eq:uplink_cost}
\end{equation}

\begin{equation}\label{eq:downlink_cost}
C_{\text{downlink}} =
\begin{cases}
\displaystyle \sum_{\ell} d_\ell \cdot 32, & \text{full precision},\\[4pt]
\displaystyle \sum_{\ell}\!\left(
  d_\ell \left\lceil \log_2 L^\downarrow_\ell \right\rceil
  + \frac{b \, L^\downarrow_\ell}{T^\downarrow_\ell}
\right), & \text{quantised}. 
\end{cases}
\end{equation}

where $b$ is the number of bits per bucket boundary, $L$ is the number of quantisation levels, and $d$ is the model dimensionality. The first term, $b \times L$, accounts for the total number of bits needed to transmit $L$ bucket boundaries (each encoded with $b$ bits). In Equation \eqref{eq:uplink_cost}, the term $\sum_\ell d_\ell\,\lceil \log_2 L_\ell\rceil$ counts \emph{indices}: each of the $d_\ell$ coordinates in layer $\ell$ uses $\lceil \log_2 L_\ell\rceil$ bits. The term $\sum_\ell \tfrac{b\,L_\ell}{T_\ell}$ accounts for the \emph{codebook}, which is transmitted only once every $T_\ell$ rounds and amortised; if endpoints are also sent, replace $L_\ell$ with $L_\ell{+}1$. For a full-precision downlink, $C_{\text{downlink}}=\sum_\ell 32\,d_\ell$; otherwise apply the same index+codebook decomposition as in Equation \eqref{eq:uplink_cost}.

\subsection{Problem Formulation}

Our objective is to minimise communication costs while preserving model accuracy. Formally:
\begin{equation}
    \min_{\mathcal{Q}} \quad C_{\text{quantized}} \quad \text{subject to} \quad \text{Acc}(\tilde{w}) \geq \text{Acc}_0 - \epsilon
\end{equation}
where $\tilde{w}$ denotes the quantized model, $\text{Acc}_0$ is baseline accuracy, and $\epsilon$ is a small tolerance. In our training process, each client optimises its local objective $F_i(w)$ using stochastic gradient descent (SGD) during local model updates.

We now introduce the proposed \textsc{QuantFL} framework and the associated quantisation methodologies in detail as depicted in Fig. \ref{fig:system_model}.

\section{Methodology: \textsc{QuantFL} Framework}
\label{sec: methodology}

In this section, we describe the proposed \textsc{QuantFL} framework, which incorporates quantised model updates to enable communication-efficient FL starting from a partially pre-trained model.

We consider cross-device FL initialised from a partially pre-trained global model. In each communication round, selected clients perform local training, compress their \emph{model updates} with a simple bucketed scalar quantiser, and send compact indices (plus infrequently refreshed boundaries) to the server, which reconstructs by mid-point decoding and aggregates. We compress \emph{updates} in every round, not the initial model.

\begin{algorithm}[t]
\caption{Federated Learning with Bucketed Update Quantisation (\textsc{QuantFL})}
\label{algo:QUANTFL}
\begin{algorithmic}[1]
\State \textbf{Server initialises} global model $w^{0}$ (partially pre-trained)
\For{round $k = 0,1,\dots,K-1$}
  \State Server samples clients $\mathcal{S}_k$ and broadcasts $w^{k}$
  \For{each client $i \in \mathcal{S}_k$ \textbf{in parallel}}
    \State Local update: train on $D_i$ to obtain $w_i^{k}$; set $\Delta w_i^k \!\gets\! w_i^{k} - w^{k}$
    \State Per-layer quantise: $(\tilde{w}_{i,k}, \{\mathcal{B}_{\ell}\}_{\text{if refresh}}) \!\gets\! Q(\Delta w_i^k)$
    \State Uplink: send indices (and boundaries if this is a refresh round)
  \EndFor
  \State Server decodes by mid-points and aggregates:
  \[
    w^{k+1} \;=\; w^{k} + \frac{1}{|\mathcal{S}_k|}\sum_{i\in\mathcal{S}_k}\tilde{w}_{i,k}
  \]
\EndFor
\end{algorithmic}
\end{algorithm}

\subsection{Bucketed Scalar Quantisation (Per-Layer)}
\label{subsec:quant}
For each layer $\ell$, we quantise the update vector over an estimated range $[m_\ell, M_\ell]$ using $L_\ell$ disjoint \emph{buckets} with boundaries $\mathcal{B}_\ell=\{b_0,\dots,b_{L_\ell}\}$, where $b_0=m_\ell$ and $b_{L_\ell}=M_\ell$. The ordered boundary list $\mathcal{B}_\ell$ is the \emph{codebook} shared by client and server. Each scalar entry $u$ is replaced by its \emph{bucket index}
$q(u)\!\in\!\{0,\dots,L_\ell\!-\!1\}$, costing $\lceil \log_2 L_\ell\rceil$ bits; the server reconstructs by the interval mid-point:
\begin{equation}
\hat{u} \;=\; \tfrac{1}{2}\big(b_j + b_{j+1}\big)\quad \text{for}\quad u\in(b_j,b_{j+1}].
\end{equation}
Mid-point decoding is MSE-optimal within an interval for scalar quantisers. Because pre-training typically reduces the dynamic range $R_\ell\!=\!M_\ell-m_\ell$, the step size $\Delta_\ell\!\approx\!R_\ell/L_\ell$ shrinks, yielding lower distortion at fixed $L_\ell$ (or the same distortion with fewer levels).

\subsection{Quantiser Instantiations}
\label{subsec:instantiations}
We study two boundary constructions for bucketed scalar quantisation and include a stochastic comparator.

\paragraph*{Bucket-Uniform (BU)}
Equal-width boundaries:
\begin{equation}
b_j \;=\; m_\ell + j\cdot\frac{M_\ell - m_\ell}{L_\ell},\qquad j=0,\dots,L_\ell.
\end{equation}
BU is simple, robust under heterogeneity, and our default unless stated.

\paragraph*{Bucket-Quantile (BQ)}
Equal-mass (empirical quantile) boundaries computed from the empirical CDF of the layer updates: each bucket contains approximately the same number of coordinates. BQ adapts to peaked distributions.

\paragraph*{Stochastic Quantisation (QSGD, comparator)}
Following~\cite{QSGD}, each coordinate is quantised probabilistically relative to its magnitude and a fixed number of levels, producing an unbiased estimate. We include QSGD as a baseline comparator rather than as part of \textsc{QuantFL}.

\subsection{Codebook Refresh and Bit Budget}
\label{subsec:refresh}
For layer $\ell$, the \emph{codebook} $\mathcal{B}_\ell$ is the ordered list of bucket boundaries that partitions the update range $[m_\ell,M_\ell]$ into $L_\ell$ intervals. We use
\[
\mathcal{B}_\ell=\{b_0,\ldots,b_{L_\ell}\},\quad b_0=m_\ell,\; b_{L_\ell}=M_\ell,
\]
so there are $L_\ell{+}1$ transmitted boundary values. During uplink, clients send only \emph{indices}; the server performs mid-point decoding with $c_j=\tfrac{b_j+b_{j+1}}{2}$. The codebook is \emph{refreshed infrequently} (every $T_\ell$ rounds) and broadcast by the server; we amortise its bit cost. For BU, $\mathcal{B}_\ell$ is determined by $m_\ell$, $M_\ell$, and $L_\ell$; for BQ, $\mathcal{B}_\ell$ is computed from early-round quantiles on the server (bootstrap) and then refreshed.

Boundaries change slowly. We therefore \emph{refresh} $\mathcal{B}_\ell$ infrequently (every $T_\ell$ rounds) and amortise their cost; per-round uplink consists primarily of indices. The per-client, per-round uplink cost is
\begin{equation}
C_{\text{uplink}}
\;=\;
\sum_{\ell}
\Big(
d_\ell \cdot \lceil \log_2 L_\ell \rceil
\;+\;
\frac{b \cdot L_\ell}{T_\ell}
\Big),
\label{eq:uplink_cost_main}
\end{equation}
where $d_\ell$ is the number of parameters in layer $\ell$ and $b$ is the precision (bits) used per transmitted boundary.

We transmit only the $L_\ell$ \emph{internal} boundaries; endpoints are recovered from $(m_\ell,M_\ell)$. The amortised codebook cost is therefore $\sum_\ell \tfrac{b\,L_\ell}{T_\ell}$; if endpoints are also sent, replace $L_\ell$ with $L_\ell{+}1$ consistently throughout.

\subsection{Implementation Details}
\label{subsec:impl}
Per-layer ranges, for BU we estimate $[m_\ell,M_\ell]$ from the pre-trained snapshot and refresh every $T_\ell$ rounds; for BQ we bootstrap quantile boundaries on the server from early-round statistics and refresh infrequently. In the default settings, unless noted, we use BU with $L_\ell=L$ for all layers and a common refresh period $T_\ell=T$.  During the aggregation phase, we use simple averaging in Algorithm~\ref{algo:QUANTFL}; dataset-size weighting can be substituted without changing the compressor.

\subsection{Convergence and Efficiency (Qualitative)}
\label{subsec:conv}
We do not optimise \eqref{eq:uplink_cost_main} directly. Instead, we select $(L_\ell,T_\ell)$ to bound quantisation distortion so that FedAvg convergence behaviour is preserved. Empirically (Section~\ref{sec:experiments}), starting from a pre-trained model concentrates updates (smaller range/variance), so bucketed mid-point quantisation incurs lower error for a given budget; this reduces the loss and total communication compared with training from scratch. BU remains stable under non-IID data.

\begin{table*}[htbp]
\caption{Training and Test Results with Communication Costs on MNIST and CIFAR-100. NQ - non-quantised, NP - not-pre-trained, Bucket-Quantile (BQ), Bucket-Uniform (BU) }
\vspace{-0.3cm}
\label{tab:performance}
\scriptsize
\centering
\begin{tabular}{lccccccc}
\toprule
\textbf{Dataset} & \textbf{Method} & \textbf{Train Accuracy} & \textbf{Train Loss} & \textbf{Test Accuracy} & \textbf{Test Loss} & \textbf{Comm. Cost (64 / 128)} & ↓ vs NQ \\ 
\midrule
\multirow{7}{*}{MNIST} 
& Baseline (NQ, NP)           & 90.37\% & 0.2760 & 90.01\% & 0.1789 & 3,494,400 &  - \\
& BQ (Ours)             & 78.84\% & 0.7964 & 77.67\% & 0.3452 & 2,074,903 / 2,129,605 & 40.62\%\ / 39.06\% \\
& BU (Ours)              & 89.18\% & 0.5376 & \textbf{89.00\%} & 0.2407 & 2,074,903 / 2,129,605 & 40.62\% / 39.06\%\\
& QSGD                        & 83.44\% & 0.6121 & 82.89\% & 0.2340 & 2,129,560 / 2,184,160 & 39.06\% / 37.50\%\\
& BU (NP)         & 70.45\% & 0.8421 & 69.78\% & 0.4020 & 2,074,903 / 2,129,605 & 40.62\% / 39.06\%\\
& BQ (NP)          & 78.12\% & 0.5910 & 77.32\% & 0.2950 & 2,074,903 / 2,129,605 & 40.62\% / 39.06\%\\
& QSGD (NP)                    & 74.54\% & 0.6480 & 73.67\% & 0.3150 & 2,129,560
 / 2,184,160 & 39.06\% / 37.50\%\\
\midrule
\multirow{7}{*}{CIFAR-100} 
& Baseline (NQ, NP)           & 66.00\% & 1.0200 & 47.00\% & 1.3100 & 5,891,001 / 6,046,446 & - \\
& BQ (Ours)             & 28.75\% & 1.9940 & 25.78\% & 2.0380 & 5,891,001 / 6,046,446 & 40.62\%\ / 39.05\% \\
& BU (Ours)             & \textbf{66.74\%} & 1.0520 & \textbf{66.89\%} & 1.2430 & 5,891,001 / 6,046,446 & 40.62\%\ / 39.05\% \\
& QSGD                        & 41.62\% & 1.6110 & 28.90\% & 1.8160 & 6,046,257 / 6,201,272 & 39.06\% / 37.49\%\\
& BQ (NP)         & 24.10\% & 2.1010 & 22.50\% & 2.2300 & 5,891,001 / 6,046,446 & 40.62\%\ / 39.05\% \\
& BU (NP)          & 60.50\% & 1.1100 & 50.30\% & 1.3100 & 5,891,001 / 6,046,446 & 40.62\%\ / 39.05\% \\
& QSGD (NP)                    & 38.40\% & 1.7000 & 26.20\% & 1.9100 & 6,046,257
 / 6,201,272 & 39.06\% / 37.49\%\\
\bottomrule
\end{tabular}
\end{table*}

\begin{table}[t]
\centering
\caption{Non-IID robustness on CIFAR-100 with Dirichlet heterogeneity (\(L{=}64\)). Smaller \(\alpha\) = stronger skew.}
\vspace{-0.3cm}
\label{tab:dirichlet}
\begin{tabular}{c|ccc}
\toprule
$\boldsymbol{\alpha}$ & \textbf{BU Acc (\%)} & \textbf{BQ Acc (\%)} & \textbf{QSGD Acc (\%)} \\
\midrule
1.0 & 68.28 & 31.0 & 31.8 \\
0.5 & 66.89 & 25.78 & 28.90 \\
0.1 & 52.1 & 23.4 & 27.1 \\
\bottomrule
\end{tabular}
\end{table}
\vspace{-0.8cm}


\begin{figure}[t]
\begin{minipage}[b]{1.0\linewidth}
  \centering
  \centerline{\includegraphics[width=8.5cm]{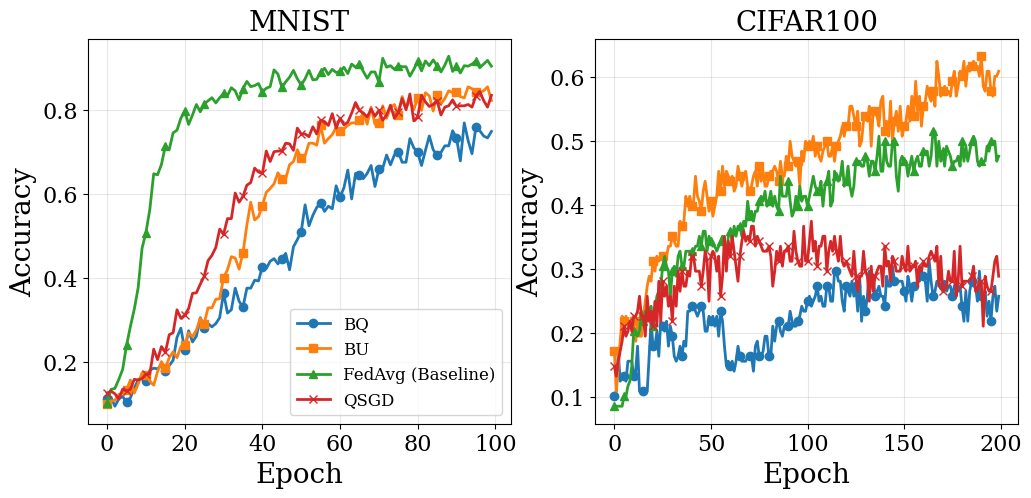}}
  \centerline{(a) Accuracy against epoch}\medskip
\end{minipage}
\begin{minipage}[b]{1.0\linewidth}
  \centering
  \centerline{\includegraphics[width=8.5cm]{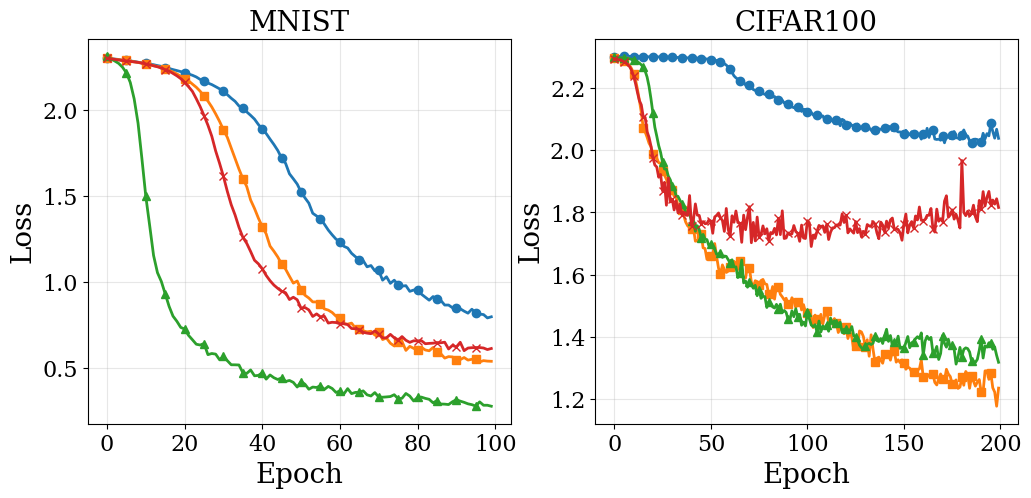}}
  \centerline{(b) Loss against epoch}\medskip
\end{minipage}
\vspace{-1cm}
\caption{Training loss and test accuracy curves for MNIST and CIFAR-100 under different quantisation methods. Baseline FedAvg is a non-quantised and non-pre-trained setting.} 
\vspace{-0.6cm}
\label{fig:res}
\end{figure}

\section{Results and Discussion}
\label{sec:experiments}

\subsection{Experimental setup}

The experiments were conducted on a high-performance computing setup, utilising an NVIDIA RTX 6000 GPU with 48GB of VRAM, coupled with an Intel Core i9-10980 processor. We evaluate our approach on two standard benchmarks: MNIST and CIFAR-100. To reflect realistic FL scenarios, we use convolutional neural networks (CNNs): a shallow CNN for MNIST and a moderately deeper CNN for CIFAR-100, which involved a more intricate model architecture (ResNet-18). ResNet-18 was chosen to accommodate the complexity of the dataset with its 100 classes. Experiments simulate a federated setup with 100 users, randomly selecting 10 clients per communication round. Each client performs two epochs of local training before transmitting updates. We compare bucket-based quantisation methods (uniform and quantile) against QSGD and a non-quantised FedAvg baseline. \emph{Quantisation:} $L\!\in\!\{64,128\}$, boundary precision $b{=}16$ bits, boundary refresh every $T$ rounds (we report $T$), and identical coding for uplink and (when used) downlink broadcast. \emph{Baselines and ablations:} Full precision FedAvg (NQ), QSGD, BU, and BQ; ablations for pre-trained vs.\ scratch initialisation and for different $L$. \emph{Metrics:} Test accuracy, uplink+downlink bits per round, and convergence behaviour across communication rounds. Table~\ref{tab:performance} summarises the training and test performance, as well as the communication costs, across all methods at two quantisation levels (64 and 128 buckets).



\subsection{Simulation Results}

\begin{figure}
    \centering
    \includegraphics[width=1\linewidth]{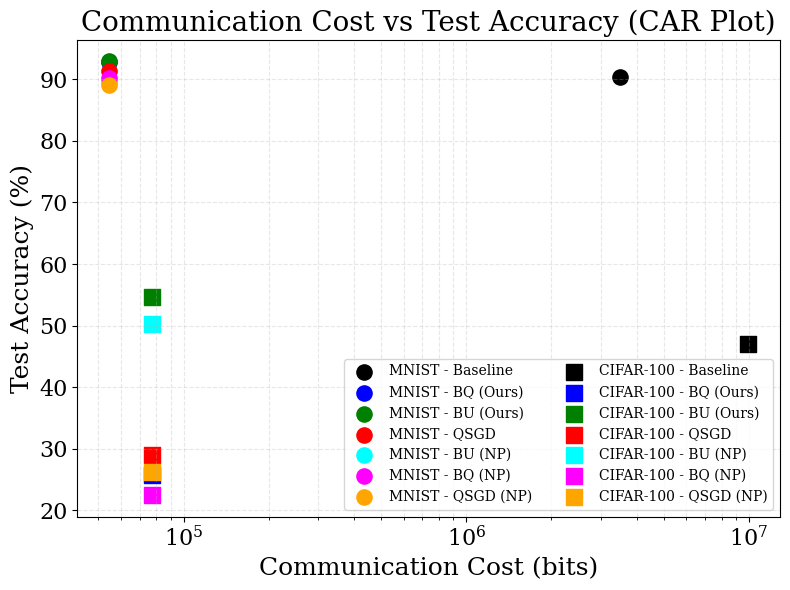}
    \vspace{-0.8cm}
    \caption{Test accuracy vs.\ total bits per round (per client; uplink+downlink; \emph{log} x-axis) on MNIST and CIFAR-100. Downlink is full-precision in all points unless stated.}
    \vspace{-0.5cm}
    \label{fig:commCost}
\end{figure}

Table~\ref{tab:performance} shows that under tight bit budgets, BU attains the highest test accuracy among quantised methods, reaching 89.00\% at \(L{=}64\) with a per-client, per-round total of 2{,}074{,}903 bits (uplink\,+\,downlink), and 2{,}129{,}605 bits at \(L{=}128\). This corresponds to 40.62\% and 39.06\% reductions versus the non-quantised (NQ) baseline, respectively. QSGD trails BU at \(\,82.89\%\) with totals of \(2{,}129{,}560\) (64) and \(2{,}184{,}160\) (128), i.e., \(\,39.06\%\) and \(\,37.50\%\) reductions. BQ is less competitive on MNIST (77.67\%), consistent with its sensitivity to distributional tails. The NP (scratch) ablations keep the same bit budgets (compression parameters unchanged) but show the expected accuracy drop (e.g., BU (NP) 69.78\%), reinforcing that pre-training improves both stability and the accuracy–bits trade-off.

On the more challenging, heterogeneous task CIFAR-100 under ResNet-18, BU remains robust, achieving a test accuracy of 66.89\% at \(L{=}64\) with 5{,}891{,}001 bits/round, and \(\,L{=}128\) with 6{,}046{,}446 bits/round 40.62\% and 39.05\% below the NQ baseline, respectively (Table~\ref{tab:performance}). BQ degrades markedly on CIFAR-100 (25.78\%), and QSGD attains \(28.90\%\) at similar budgets, indicating that a uniform bucket allocation is more tolerant to heavy-tailed, non-IID update distributions. The NP counterparts again underperform their pre-trained versions (e.g., BU (NP) 50.30\%), showing that pre-training systematically improves final accuracy for a fixed communication budget.

Table~\ref{tab:dirichlet} reports a Dirichlet heterogeneity sweep at \(L{=}64\). As skew increases (\(\alpha \downarrow\)), all methods degrade, but BU remains clearly ahead (\(68.28\%\) at \(\alpha{=}1.0\), \(66.89\%\) at \(\alpha{=}0.5\), \(52.1\%\) at \(\alpha{=}0.1\)), while BQ and QSGD remain substantially lower. This aligns with the design intuition: BU’s equal-width buckets are robust when update distributions have pronounced tails, whereas BQ’s equal-mass buckets are more brittle under skew.

Overall, (i) Pre-training helps: for a fixed bit budget, pre-trained initialisation consistently yields higher accuracy than scratch (NP) variants on both datasets. (ii) BU is a strong default: it dominates BQ and QSGD on CIFAR-100 and is best among compressed methods on MNIST. (iii) Energy-Accuracy Trade-offs: While complex sparsification methods exist, they often degrade performance under non-IID data, a typical scenario in IoT sensor networks. 

As seen in the Fig.~\ref{fig:commCost}, \textsc{QuantFL} sits on the Pareto frontier of accuracy vs.~cost. Although our total reduction is $\approx 40\%$ (due to the strategic choice of preserving a robust full-precision downlink), the uplink cost, which dominates the energy consumption of IoT transmitters is reduced by orders of magnitude (from 32 bits to $\approx 6$ bits per coordinate). This makes \textsc{QuantFL} uniquely suited for sustainable deployments where uplink battery life is the primary bottleneck.

Thus, \textsc{QuantFL} demonstrates that signal processing-inspired structured quantisation, combined with partial pre-training, enables scalable and efficient FL suitable for bandwidth-constrained edge IoT environments.

\section{Conclusion}


We introduced \textsc{QuantFL}, a sustainable FL framework designed for energy-constrained Edge IoT. By starting from a pre-trained model and quantising per-layer client updates every round using simple bucketed scalar quantisers. The core observation is that pre-training concentrates updates, reducing their dynamic range and variance. So a small number of code levels with mid-point decoding yields low distortion. In our design, clients transmit only bucket indices each round, while codebooks (boundary lists) are refreshed infrequently and their cost is amortised; the bit budget explicitly separates index bits from codebook overhead. Across MNIST and CIFAR-100, \textsc{QuantFL} delivers large uplink savings while maintaining competitive accuracy. The approach is simple to implement, works with standard FedAvg aggregation, and is orthogonal to other compressors (e.g., sparsification, low-rank) and to potential downlink compression. However, our evaluation focuses on image benchmarks and moderate backbones as future works; extending to larger models/datasets and more realistic cross-device settings (partial participation dynamics, stragglers, energy constraints) is a priority. An adaptive policy for $(L_\ell, T_\ell)$ and codebook refresh, formal analysis of pre-training induced concentration with (biased) compressors and error feedback, and quantising the downlink are promising directions. Finally, combining \textsc{QuantFL} with personalisation and foundation-model starts may further improve both accuracy and communication.

\bibliographystyle{IEEEtran}
\bibliography{refs}

\end{document}